%% file: submission-v3.tex
\documentclass[twoside]{article}

\usepackage{aistats2019}

\input{math_commands.tex}

\usepackage{hyperref}
\usepackage{url}
\usepackage[utf8]{inputenc} 
\usepackage[T1]{fontenc}    
\usepackage{booktabs}       
\usepackage{amsfonts}       
\usepackage{nicefrac}       
\usepackage{microtype}      
\usepackage{algorithm}
\usepackage{algorithmic}
\usepackage[round]{natbib}
\usepackage{microtype}
\usepackage{graphicx}
\usepackage{subfigure}
\usepackage{amsmath}
\usepackage{booktabs}
\usepackage{color}
\usepackage{IEEEtrantools}
\usepackage{wrapfig}
\usepackage[titletoc,title]{appendix}

\newcommand\ie {{\it i.e., }}
\newcommand\eg {{\it e.g., }}
\newcommand\etc{{\it etc.}}

\newcommand\eq {{\it Eq.}}

\newcommand\DNF {{DDNF}}

\hypersetup{draft}

\begin{document}

\twocolumn[
\aistatstitle{Deep Diffeomorphic Normalizing Flows}
\aistatsauthor{Hadi Salman\\
            Microsoft Research AI \\
            \texttt{hasalman@microsoft.com} \And 
            Payman Yadollahpour\\
            University of Pittsburgh\\
            \texttt{payman@pitt.edu}\\ \And
            Tom Fletcher\\
            Universrity of Virginia \\
            \texttt{ptf8v@virginia.edu}\And
            Kayhan Batmanghelich \\
            University of Pittsburgh\\
            \texttt{kayhan@pitt.edu}
            }
            \aistatsaddress{}
          
]
\begin{abstract}
\input{abstract.tex}
\end{abstract}

\input{intro.tex}

\input{bkg.tex}

\input{ourMethod.tex}

\input{relatedWork.tex}

\input{experiments2.tex}

\input{conclusion.tex}

\bibliographystyle{plainnat}
\bibliography{Mendeley_ICLR2018,iclr2019_conference}

\cleardoublepage
\input{appendix.tex}

\end{document}

%% file: math_commands.tex
\usepackage{amsmath,amsfonts,bm}

\def\eqref#1{eq.~\ref{#1}}

\def\1{\bm{1}}

\def\rmI{{\mathbf{I}}}

\def\rmL{{\mathbf{L}}}

\def\ermX{{\textnormal{X}}}

\def\vu{{\bm{u}}}
\def\vv{{\bm{v}}}
\def\vw{{\bm{w}}}
\def\vx{{\bm{x}}}

\def\vz{{\bm{z}}}

\DeclareMathAlphabet{\mathsfit}{\encodingdefault}{\sfdefault}{m}{sl}
\SetMathAlphabet{\mathsfit}{bold}{\encodingdefault}{\sfdefault}{bx}{n}

\def\gF{{\mathcal{F}}}

\def\gJ{{\mathcal{J}}}

\def\gN{{\mathcal{N}}}
\def\gO{{\mathcal{O}}}

\def\gR{{\mathcal{R}}}

\newcommand{\E}{\mathbb{E}}



%% file: abstract.tex
The Normalizing Flow (NF) models a general probability density by estimating an invertible transformation applied on samples drawn from a known distribution. 
We introduce a new type of NF, called Deep Diffeomorphic Normalizing Flow (\DNF). A diffeomorphic flow is an invertible function where both the function and its inverse are smooth. 
We construct the flow using an ordinary differential equation (ODE) governed by a time-varying smooth vector field.
We use a neural network to parametrize the smooth vector field and a recursive neural network (RNN) for approximating the solution of the ODE. Each cell in the RNN is a residual network implementing one Euler integration step.
The architecture of our flow enables efficient likelihood evaluation, straightforward flow inversion, and results in highly flexible density estimation.
An end-to-end trained \DNF~ achieves competitive results with state-of-the-art methods on a suite of density estimation and variational inference tasks. Finally, our method brings concepts from Riemannian geometry that, we believe, can open a new research direction for neural density estimation.


%% file: intro.tex
\section{Introduction}
\label{sec:intro}

Efficient computation of the posterior distribution is one of the main problems in Bayesian Inference. The exact form of the posterior density function requires the estimation of the marginal likelihood which is computationally prohibitive in general~\citep{Valiant1979}. To approximate the posterior distribution, there are, arguably, two families of approaches: (1) methods based on sampling (\eg MCMC~\citep{Metropolis1953, Hastings1970}, Gibbs~\citep{Geman1984}, \etc), and (2) variational inference (VI) techniques~\citep{Jordan1999}. The general idea of a sampling method is to construct an ergodic chain of the latent variable sampled from the posterior. Although MCMC methods provide asymptotic guarantees for producing exact samples from the posterior~\citep{Robert2004}, they tend to be computationally expensive for large datasets or complex models.

Instead of sampling, VI techniques convert the approximation problem into an optimization problem. They maximize a lower bound that indirectly minimizes the Kullback-Leibler (KL) divergence between the exact posterior and a member of a postulated family of probability density functions~\citep{Jordan1999,Bishop2006}. Although no asymptotic guarantee is known for VI, it tends to scale better than MCMC thanks to powerful optimization techniques such as stochastic gradient descent~\citep{Robbins1951a}. The choice of the family of the distribution is important, and a not rich enough family can result in a biased approximation of the posterior~\citep{Turner2011}.

In recent years, various neural density estimators have been proposed~\citep{Mnih2014,rezende2015variational,kingma2016improved,Larochelle2011, papamakarios2017masked,huang2018neural}. These methods use neural networks to specify flexible families of distributions for VI. The challenge is to ensure that the approximate densities are easy to sample from and to evaluate.  For example \cite{rezende2015variational}   apply a series of invertible transformations on a random variable drawn from a fixed distribution (\eg a Gaussian distribution) to represent complex distributions. \cite{Larochelle2011} use an autoregressive approach which views the approximate posterior as a decomposition of a chain of conditional distributions. \cite{kingma2016improved} show that those approaches are closely related.

We propose a novel normalized flow method where the invertible function is a diffeomorphism, dubbed \DNF. A diffeomorphism is an invertible mapping where both the function and its inverse are smooth.  Inspired by the literature of large deformation diffeomorphic metric mapping in medical image registration~\citep{Beg2005,Beg2005a,younes2010shapes,Zhang2015}, we use an ODE to construct such a mapping. We propose to use an RNN to discretize the ODE where the RNN cell has a residual neural network (ResNet)~\citep{He2015} architecture.
The resulting flow can be viewed as a composition of tiny mappings that are close to the identity transformation. Generalizing some previous methods~\citep{rezende2015variational,jankowiak2018pathwise}, \DNF~is highly flexible and easy to sample from. The construction of the inverse flow,  required for evaluating the likelihood of given data samples at test time, is expressive and straightforward.
We draw a connection between neural density estimation and the Riemannian geometric structure of the manifold of diffeomorphic functions which we believe can potentially open new directions of research.\footnote{Upon publication, we will open source the code.}


%% file: bkg.tex
\section{Background}
 Given a dataset $\ermX = \{ \vx_1, \cdots, \vx_N \}$, the maximum likelihood principle is typically used to learn the parameters $\theta$ of a model given its probability distribution:
 \begin{IEEEeqnarray*}{c}
 \log p_{\theta}(\ermX) = \sum_{i=1}^{N}\log p_{\theta}(\vx_i).
 \end{IEEEeqnarray*}
Unfortunately, maximum likelihood estimation is computationally expensive in the presence of latent variable $\vz$ because evaluating the objective function entails marginalizing out $\vz$, \ie $p_{\theta}(\vx) = \int  p_{\theta}(\vz,\vx) d\vz$, which is not tractable in general. Instead, VI~\citep{Jordan1999} maximizes a lower bound constructed by an approximation of the posterior, $q_{\lambda}(\vz|\vx)$,
\begin{IEEEeqnarray*}{c}
\log p_{\theta}(\vx) \geq \underbrace{ \mathbb{E}_{q_{\lambda}(\vz | \vx)} \left[ \log p_{\theta}(\vx | \vz) - \text{KL}( q_{\lambda}(\vz | \vx) | p(\vz) ) \right]}_{\gF(\theta,\lambda)}.
\vspace{-.1cm}
\end{IEEEeqnarray*}
The bound is called the \textit{Evidence Lower Bound} (ELBO) which is a unified cost function $\theta$ and the parameters of the approximate posterior $\lambda$. The choice of $q_{\lambda}(\vz | \vx)$ is crucial, and a ``not complex enough'' distribution can result in a biased estimation of $\theta$~\cite{Turner2011}.  Using a neural network to parameterize $q_{\lambda}(\vz | \vx)$ has proven to be successful and, with the advent of variational auto-encoders (VAEs)~\citep{kingma2013auto}, has resulted in a new direction of research. The VAE models $q_{\lambda}(\vz | \vx)$ as a Gaussian distribution with diagonal covariance, \ie $q_{\lambda}(\vz | \vx) = \gN(\vz | \mu(\vx), \text{diag}(\sigma^2(\vx)) )$ where the mean and variance are a non-linear function of $\vx$. However, the typically used family of the approximate distribution is limited since it represents a uni-modal distribution given $\vx$. \cite{rezende2015variational} proposed more flexible family of distributions. The idea is to model the posterior as a series of invertible transformations $\phi_k$ applied to random variables drawn from an initial distributions $q_{0}$. A neural network is used to parameterize the transformations. Assuming $\vz_{0}\sim q_0(\vz | \vx)$ and $\vz_{1} = \phi_1(\vz_0)$, the distribution of $\vz_{1} \sim q_{1}(\vz)$ , follows,
\begin{align}
       q_1(\vz_1) &= q_0(\vz_0) \left| \det \frac{  \partial \phi_1 (\vz_0) }{ \partial \vz_0  } \right|^{-1} \nonumber \\
                  &= q_0( \phi_1^{-1} (\vz_1)) \left| \det \frac{  \partial \phi_1^{-1} (\vz_1) }{ \partial \vz_1  } \right|.
\end{align}
After applying $K$ transformations to the latent variable, \ie $z_{K} = \phi_{K} \circ \cdots \circ \phi_{2} \circ \phi_1(\vz_0)$, the variational objective can be written as:
\begin{align}
\gF(\theta,\lambda) &= \E_{q_\lambda}[ \log q_{\lambda}(\vz|\vx) - \log p_{\theta}(\vx,\vz)  ] \nonumber \\
                    &=\E_{q_0} [ \log q_{0}(\vz_0|\vx) - \log p_{\theta}(\vx,\vz_K)   ] \nonumber \\
                    &-\E_{q_0} \left[ \sum_{k=1}^K \log \left| \det \left( \frac{  \partial \phi_k (\vz_{k-1}) }{ \partial \vz_{k-1}  } \right) \right|  \right].
\end{align}
The main challenge is to design a neural network architecture for $\phi_{k}$ that is invertible and whose determinant of the Jacobian is easy to compute. \cite{rezende2015variational} proposed a family of such transformations named planar normalizing flows: $\vz_{1} = \vz_{0} + \vu h(\vw^T \vz + b)$. The function is invertible under a simple constraint and has a closed-form determinant of the Jacobian.

The planar transformation is a single layer neural network and has a limited capacity. Recently, several other methods were introduced to increase the flexibility of the flow while maintaining the invertibility~\citep{van2018sylvester,huang2018neural,tomczak2016improving} (see Section~\ref{sec:related} for more details), and many novel architectures are proposed to keep the computation of the determinant of the Jacobian tractable. 
In this paper, we propose to use a special class of invertible mapping called \emph{diffeomorphisms} which has many appealing mathematical properties. 



%% file: ourMethod.tex
\section{Diffeomorphic Normalizing Flow}
In this paper, we enforce the $\phi$ mapping to be \emph{diffeomorphic}, meaning $\phi^{-1}$ exists and both $\phi$ and $\phi^{-1}$ are differentiable. In Section~\ref{sec:DiffBkg}, we provide some background on the space of diffeomorphisms and how they can be expressed by ordinary differential equations (ODEs). In Section~\ref{sec:NNP}, we introduce a neural network implementation of the diffeomorphic flow (\DNF), and  we present efficient implementations of the determinant of the Jacobian and the inverse of the flow. Finally, in Section~\ref{sec:regularizing_the_flow}, we propose a few regularization methods that improve the performance of \DNF.

\begin{figure}[th]
  \centering
  \includegraphics[width=0.35\textwidth]{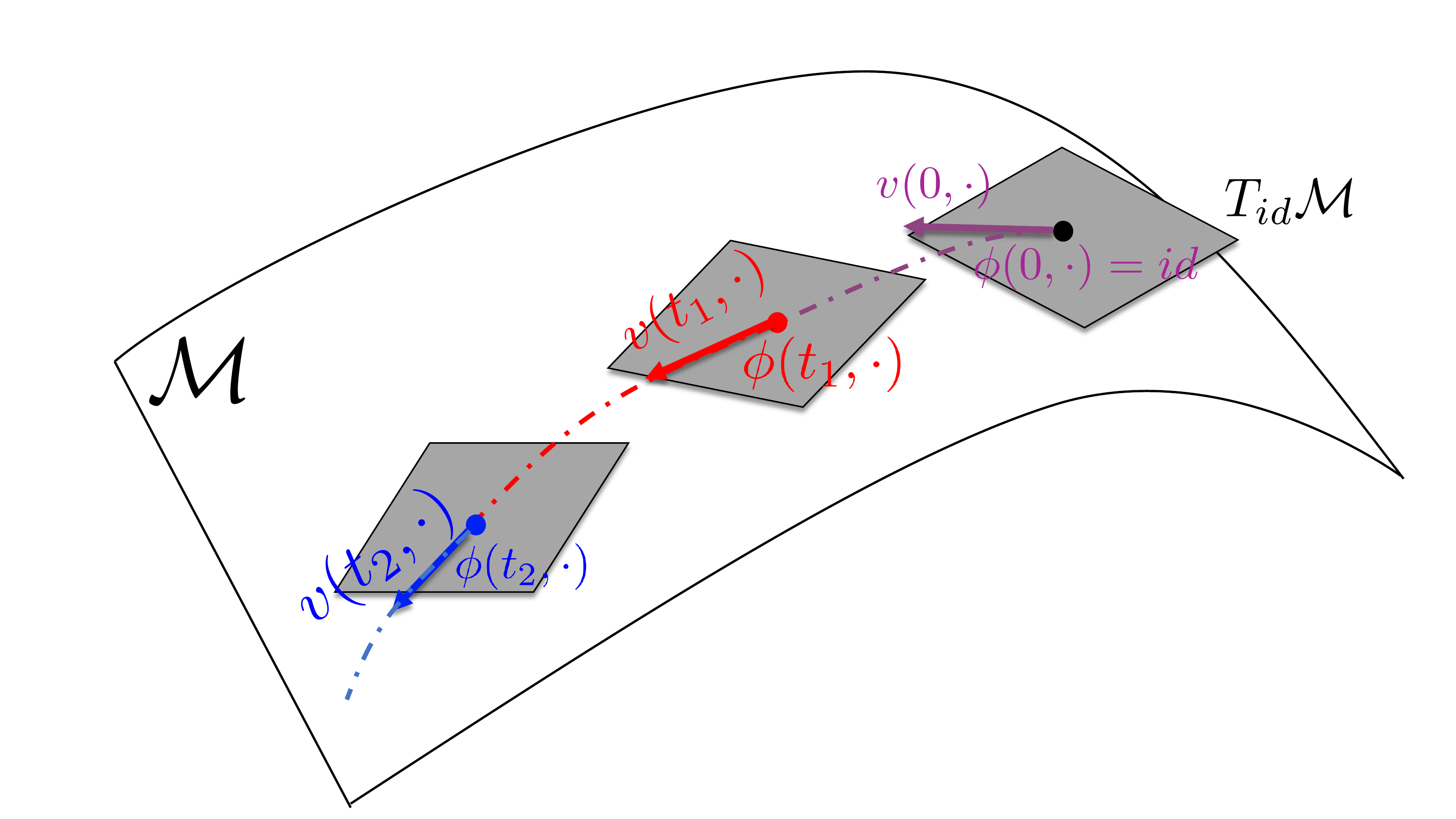}
  \caption{\small
  Representation of the non-stationary flow on the manifold of the diffeomorphic mappings, denoted $\mathcal{M}$. $\phi(t_1,\cdot)$ is the results of integrating an ODE governed by the velocity field $\vv(0,\cdot)$ in the interval of $[0,t_1]$. The velocity field belongs to the tangent space $\mathcal{T}_{id} \mathcal{M}$ at the identity transformation.  A non-stationary velocity field can be viewed as concatenation of a series of stationary flows which lead to the following composition of transformations $\phi(t_K,\cdot) \cdots \circ \phi(t_1,\cdot) \circ \phi(0,\cdot)$.
  }
  \label{fig:gen-man}
\end{figure}

\begin{figure}[th]%
    \centering
    \includegraphics[height=8cm]{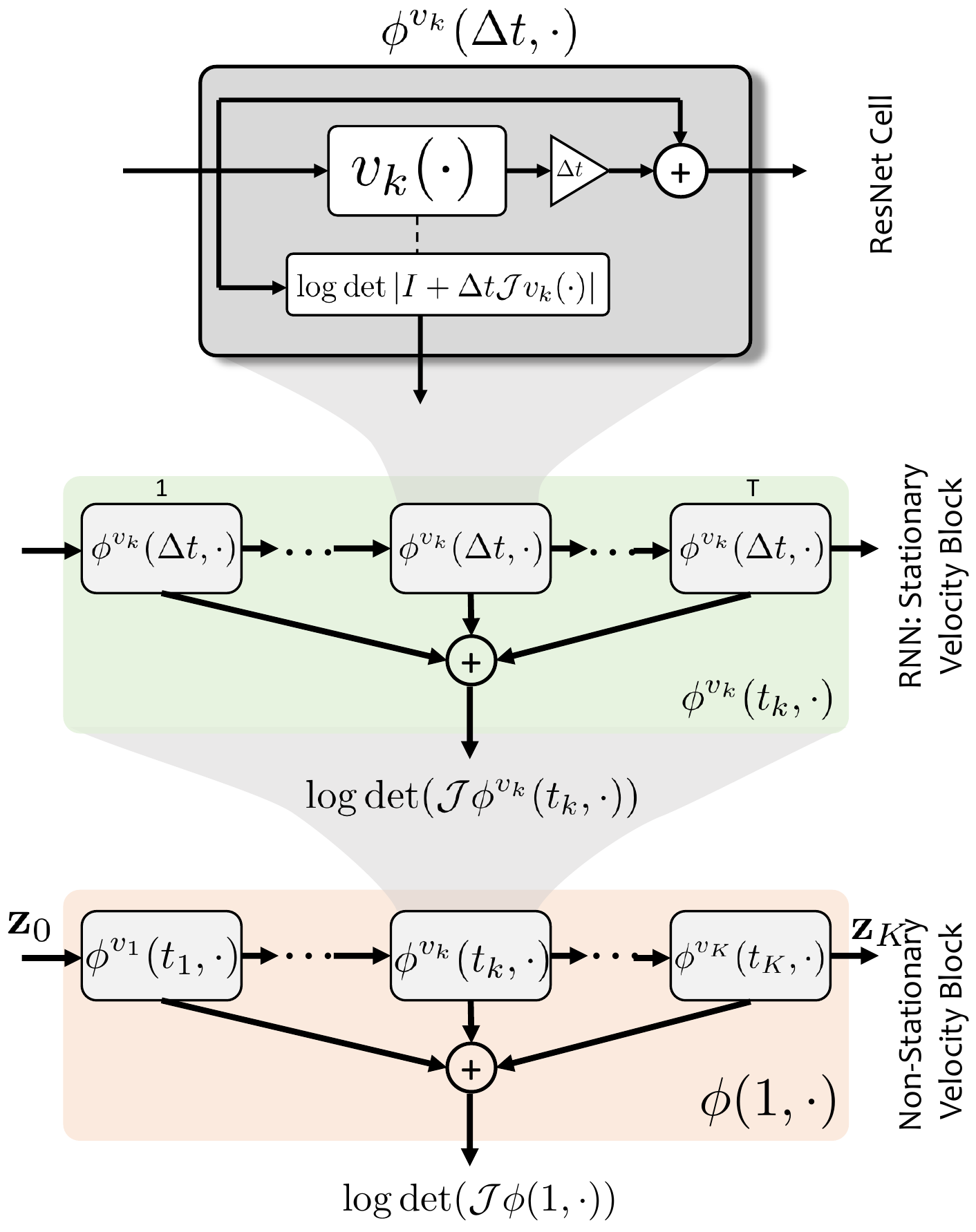} 
     \caption{\small
     General architecture of the flow. \textbf{Bottom:} The flow ($\phi(1,\cdot)$) is represented as composition of several stationary flows over time segments: $\phi(1,\cdot) = \phi^{\vv_T}(t_T, \cdot) \circ \cdots \circ \phi^{\vv_1}(t_1, \cdot)$. The velocity field for each stationary flow is different. \textbf{Middle:} The stationary flow is modeled as a composition of small flows $\phi^{\vv_i}(\Delta t, \cdot)$ sharing the velocity field. \textbf{Top:} Each of the small flows are modeled by a ResNet. The triangle simply means multiplication with $\Delta t$ and we use neural network as a general function approximator for $\vv_{i}(\cdot)$. 
     }
    \label{fig:general}%
\end{figure}

\subsection{Background on Diffeomorphisms}
\label{sec:DiffBkg}
In this section, we briefly review the mathematical background for diffeomorphic transformations. Throughout this paper, $\phi(\cdot):\Omega \rightarrow \Omega $ denotes a mapping defined on a domain $\Omega \subset \mathbb{R}^d$. 
We use $V := H^{s}(\mathcal{T} \Omega)$ to denote the Hilbert space of vector fields on $\Omega$ whose derivatives up to order $s$ exist and are square-integrable, and where $\mathcal{T} \Omega$ denotes the tangent bundle of $\Omega$. 
In this paper, we consider diffeomorphisms generated by the flow of time-varying \emph{velocity} fields. More specifically, given a time-varying vector field $\vv(t,\cdot): [0,1] \rightarrow V$, we define the time-varying flow, $\phi(t,\cdot)$, as a solution of the following differential equation:
\begin{equation}
     \tfrac{d}{dt}\phi(t,\vz) = \vv(t,\phi(t,\vz)), 
     \label{eq:ODE}
\end{equation}
where $\vv(t,\cdot)$ is a smooth function defining a velocity vector at time $t$ over its domain.
As shown in~\cite{trouve1995}, integration up to unit time (\ie $\phi(1,\cdot)$) results in a diffeomorphism if $\vv(t,\cdot)$ is sufficiently smooth
~\citep{Rossmann2002,Beg2005a,Beg2005,Hauser2017,younes2010shapes}. 
Furthermore, the determinant of the Jacobian of this diffeomorphic flow is guaranteed to be always non-negative~\citep{Gordon1972}.

We represent the time-varying velocity field, $\vv(t,\cdot)$, by segments of \emph{stationary velocity fields}, meaning that the velocity is time invariant within each segment. Hence, the overall flow is a composition of the flows governed by stationary fields. The idea is show in \figurename~{\ref{fig:gen-man}}.

The space of diffeomorphic transformations ($\mathcal{M}_d$) has several appealing properties: (1) it forms an algebraic group that is closed under the composition operation (\ie if $\phi_1, \phi_2 \in \mathcal{M}_d$ then $\phi_1 \circ \phi_2 \in \mathcal{M}_d$)~\citep{Hauser2017}, (2) with a proper definition of local inner product, all diffeomorphic transformations reside on a Riemannian manifold~\citep{younes2010shapes}. The second property allows us to define a distance metric and a notion of shortest path between two flows on the manifold. We use notion of the the shortest path as a natural regularization technique of \DNF.

In the following sections, we introduce a sub-group of diffeomorphisms defined by stationary velocity fields (Section~\ref{sec:vel_stat}). Then we extend this family to non-stationary velocity fields in Section~\ref{sec:vel_nonStat}. The general idea is shown in \figurename~{\ref{fig:general}}.

\subsection{Neural Network Parametrization}
\label{sec:NNP}

\subsubsection{Stationary Velocity Field}
\label{sec:vel_stat}
In this section, we restrict the diffeomorphisms to a special class where the velocity field in \eq~\ref{eq:ODE} is time independent (\ie $\vv(t,\vx) = \vv(\vx)$). Such restriction, also applied by~\cite{Arsigny2006,arsigny2006processing,Vercauteren2007b}, defines a subgroup of diffeomorphisms governed by the stationary ODE,
\begin{equation}
     \tfrac{d}{dt}\phi^\vv(t,\vz) = \vv (\phi(t,\vz)). 
     \label{eq:ODE-stationary}
\end{equation}
The solution of this ODE is the exponential map of the vector field, \ie $\phi^{\vv}(1,\cdot)= Exp \left( \vv(\cdot) \right)$. To compute the exponential map, we adopt an Euler integration approach that composes infinitesimal flow fields successively. In other words, the exponential map can be viewed as a composition of $T$ small flows, $\phi^{\vv}(1,\cdot)=Exp \left( v(\cdot)/T \right)^{T}$. Discretizing \eq~\ref{eq:ODE-stationary} over time, we arrive at,
\begin{equation}
    \phi^{\vv}({t + \Delta t}, \vz) = \phi(t,\vz) + \Delta t \times \vv(\phi(t,\vz)).
    \label{eq:odeDisc}
\end{equation} 
We use an RNN to model \eq~\ref{eq:odeDisc} as shown in~\figurename~{\ref{fig:general}}-Middle. Each cell has a ResNet architecture as shown in~\figurename~{\ref{fig:general}}-Top. In order to set $\Delta t = \tfrac{1}{T}$, the RNN should be unfolded $T$ times. We use a deep neural network as a general parameterization for $\vv(\cdot)$ inside ResNet.

\subsubsection{Computation of Determinant Jacobian}
Our \DNF~ applies a series of transformations to a random variable. In order to compute the probability density function of the transformed random variable, the computation of the determinant of the Jacobian ($\mathcal{J}$) of each transformation is required. 
The computational complexity of the determinant of the Jacobian of the entire flow is $\mathcal{O}(Td^3)$ where $T$ is the number of cells in the flow and $d$ is the dimension of the vector field. 
Each cell of \DNF~applies a small transformation to the random variable. In other words, for $\Delta t = \tfrac{1}{T}$, $\phi^{\vv}(1,\cdot)$ can be viewed as 
\begin{equation}
\phi^{\vv}(1,\cdot) = \underbrace{ \phi^{\vv}(\Delta t, \cdot) \circ \cdots \circ \phi^{\vv}(\Delta t, \cdot)  }_{T}. 
\end{equation}
Each $\phi^{\vv}(\Delta t, \cdot)$ is a identity-like transformation; hence we can use the Taylor series expansion around the identity to approximate the determinant of the Jacobian (see Appendix~D for derivation) as follows,
\begin{align}\label{eq:taylor}
   & \log \det(\gJ \phi^{\vv}(\Delta t,\cdot)) = \left( I+\Delta t \mathcal{J} \vv(\cdot) \right)  \\
   &\approx \tfrac{1}{2}\Delta t \text{Tr}(\gJ \vv(\cdot) + \gJ \vv(\cdot)^T) - \tfrac{1}{2}(\Delta t)^2 \text{Tr}(\gJ \vv(\cdot)^T \gJ \vv(\cdot)). \nonumber
\end{align}
A naive storage cost of the trace is $\gO(d^2)$, but the cost can be reduced by a randomized method~\citep{Hutchinson1990,EarlyStopping} as follows,
{\small
\begin{IEEEeqnarray}{lll}
  \text{Tr}(\gJ \vv(\cdot)) \approx  \frac{1}{M} \sum_{m=1}^M \vw_m^T \gJ \vv(\cdot) \vw_m, & \quad & \vw_m \sim \mathcal{N}(0,I),
  \label{eq:rndTrace}
\end{IEEEeqnarray}
}
which requires efficient Jacobian vector computation resulting in a cost reduction of the original determinant from $\gO(d^3)$ to $\gO(Md)$.

\subsubsection{Inversion of the Flow}
Invertibility of the flow enables us to evaluate the posterior distribution of any given latent variable $\vz$. 
Not all NFs has straightforward inversion. Previous approaches, such as the planar NF~\citep{rezende2015variational}, construct each cell to be invertible by imposing constraints on the parameters of the neural network. However, we construct our flow as an exponential map with no constraint on the neural network. In our approach, the inverse flow is obtained by integrating the negative velocity field in time~\citep{Rossmann2002}, \ie $\phi^{-1}(1,\cdot) = \text{Exp}(-\vv(\cdot))$. In other words, the inverse flow is another \DNF~implementing the ODE with $-\vv(\cdot)$. As the number of cells increases (\ie~$\Delta t \rightarrow 0 $), the accuracy of the approximate ODE also increases, resulting in a more accurate inversion of the flow.

\subsubsection{Extension to the Time-Varying Velocity Field}
\label{sec:vel_nonStat}
The stationary velocity field in Section~\ref{sec:vel_stat} is implemented as a series of $T$ ResNet cells sharing the same parameters composing one RNN block. Extending the method to time-varying velocity fields is straightforward; we divide the unit interval $[0,1]$ into $K$ segments. For each segment, we use a stationary velocity block with a different set of parameters. Therefore, the resulting architecture is a non-stationary velocity block as shown in Figure~\ref{fig:general}-Bottom.

\subsection{Regularizing the Flow}
\label{sec:regularizing_the_flow}
The structure of our flow suggests two interesting regularizations that improve its performance and stability:  (1) geodesic regularization and (2) an inverse consistency regularization. 

\subsubsection{Geodesic regularization}
The $\phi(1,\cdot)$ is the final point of a path defined by the ODE which is parametrized by a time-varying $\vv \in L^{2}(V,[0,1])$. There are infinite paths ($\vv$'s) connecting $id$ to $\phi(1,\cdot)$. The length of the path indicates a distance of $\phi(1,\cdot)$ from the identity mapping (\ie $id$) on $\mathcal{M}$. One may define the optimal velocity field as being the one with minimum path-length, $\Gamma(\vv)$,  defined as,
\begin{equation}
\Gamma(\vv) = \int_{0}^{1}{ \| \vv(t,\cdot) \|_{V}^{2} dt},
\label{eq:geodes}
\end{equation}
where $\| \vv(t,\cdot) \|_{V}^{2}$ is a Hilbert norm.  We define $\| \vv(t,\cdot) \|_{V}^{2}$ using the inner product on the space of velocities, $V$:
 \begin{equation}
   \left< \vv(\vz),\vw(\vz) \right>_{V} = \mathbb{E}_{q_0} \left[ v(\vz)^T \rmL w(\vz) \right],
 \end{equation}
 where $\rmL$ is a positive definite operator. In this paper, we simply choose $\rmL$ to be the  identity operator. To implement the integral in~\eq~\ref{eq:geodes}, we use the sum of the $\ell_2$-norm of the velocity vectors of cells as a regularizer (see the Appendix~E for more choices of the regularizer). 

\begin{figure}[t]
\centering
\includegraphics[height=3.4cm]{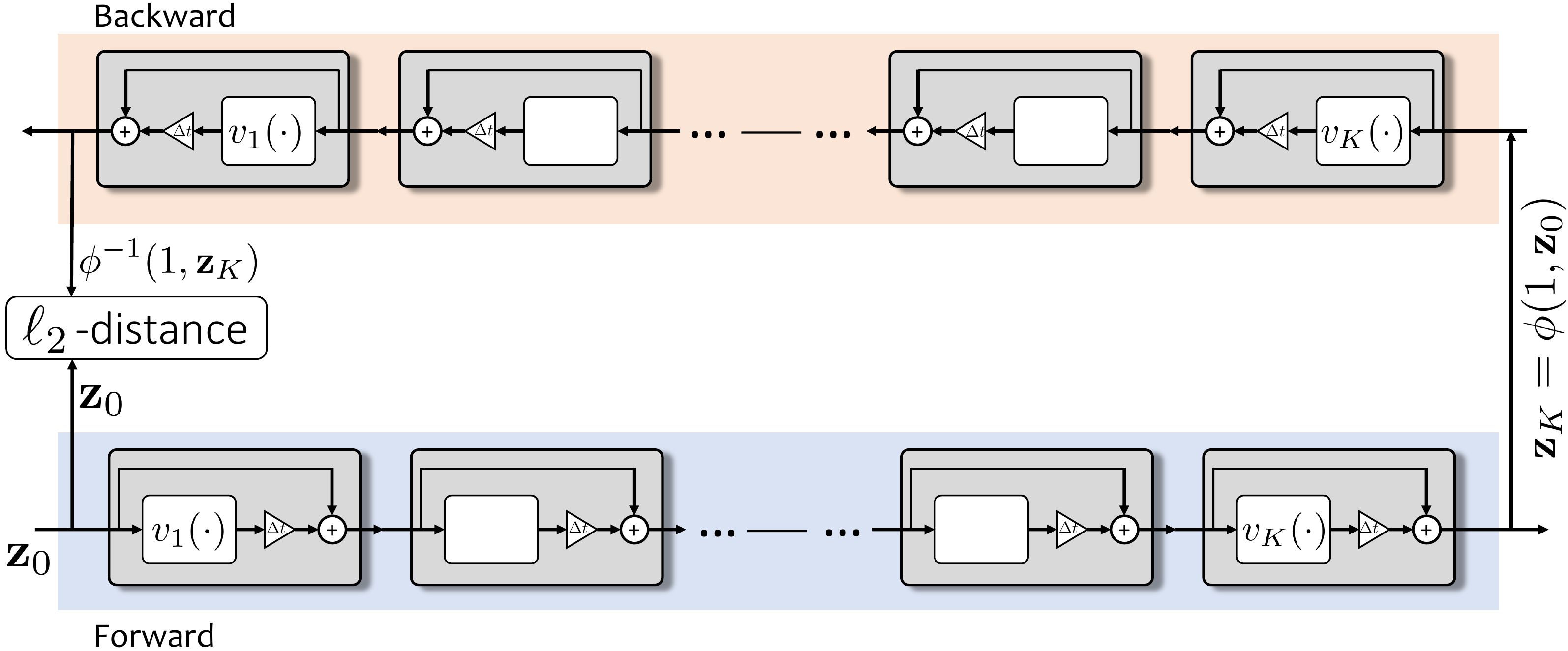}
 \vspace{-0.7cm}
\caption{\small 
Implementation of the inverse consistency regularizater using the forward and backward flows. $\vz_{K}$ is the result of passing  $\vz_{0}$ through the flow. The regularizer enforces $ \vz_0 \approx \phi^{-1}(1,\vz_K)$. }
\label{fig:invConst}
\end{figure}

\begin{figure*}[t]
   \centering
   \includegraphics[height=4cm]{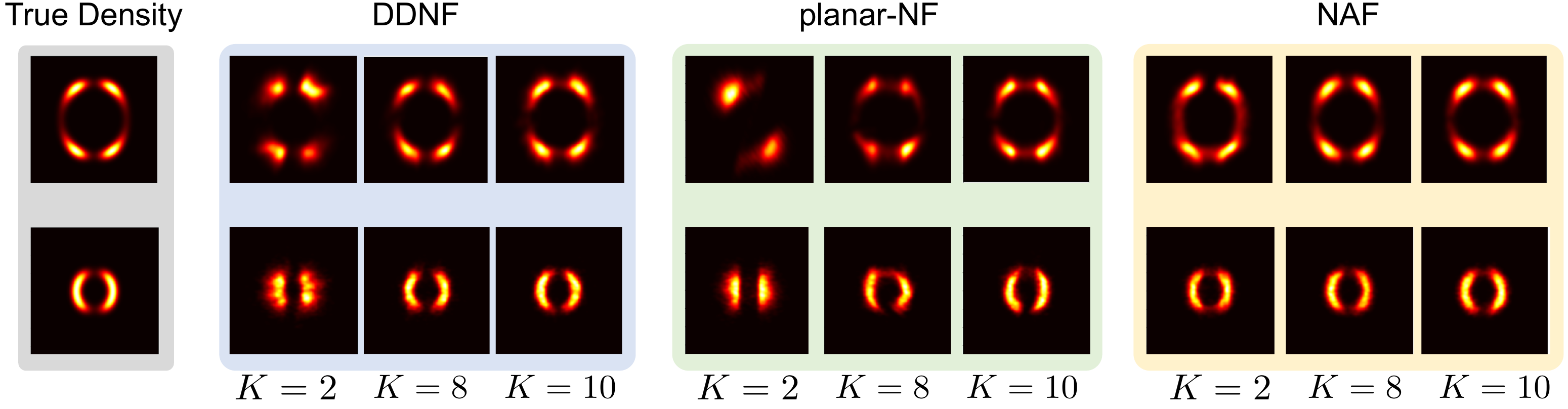}
   \vspace{-0.5cm}
   \caption{\small Comparing three flow methods in representing two toy distributions.}
   \label{fig:toyDensity}
\end{figure*}

\begin{figure}[t]
   \centering
   \includegraphics[width=.8\linewidth]{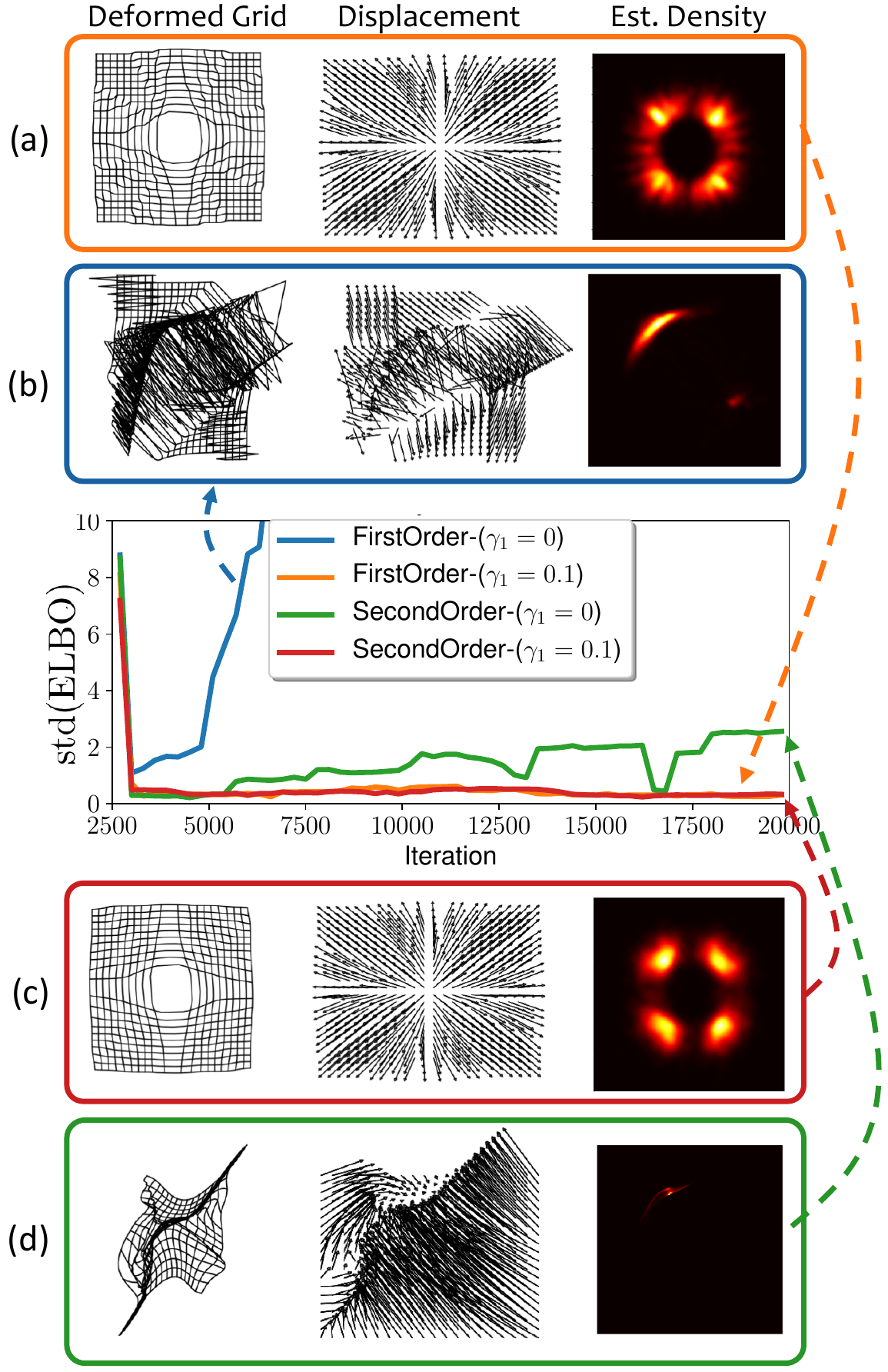}
   \caption{\small
   Stability analysis of \DNF~with $T=1$ for various regularization scenarios. The deformation grid, displacement field, and the estimated density are shown for \texttt{FirstOrder} (a) with regularization and (b) without regularization, and for \texttt{SecondOrder} (c) with regularization and (d) without regularization.  \texttt{FirstOrder} and \texttt{SecondOrder} denote the linear and the quadratic approximation (wrt to $\Delta t$) of the $\log \det(\cdot)$ in~\eqref{eq:taylor}. The stability of each of the setups is measured by reporting the standard deviation of ELBO during training.}
   \label{fig:stability}
\end{figure}
 
\subsubsection{Inverse Consistency Regularization}
We use Euler discretization for the integration of the ODE in \eq~\ref{eq:ODE}. The quality of flow inversion increases with the number of cells (\ie $\Delta t \rightarrow 0$). However, a large number of cells increases the computational cost for the forward and the backward passes. The geodesic regularization (\eq~\ref{eq:geodes}) improves both the quality of the ODE and $\log \det (\cdot)$ approximations but a strong regularization of the velocity field results in a stiff flow (\ie $\phi \approx id$). To improve the quality of the inversion, we propose an inverse consistency regularization. For an invertible flow, $\vz_0 = \phi^{-1}(1,\vz_K)$
where $\vz_{0}$ is the sample from the initial distribution and $\vz_{K}$ is the final output of the forward flow (\ie $\vz_{K} = \phi(1,\vz_0)$). The proposed regularization enforces them to be close,
\begin{equation}  \label{eq:invConst}
  \gR(\vv) = \|  \vz_0 - \phi^{-1}(1,\vz_K) \|_2.
\end{equation}
The general idea is shown in \figurename~{\ref{fig:invConst}}.

%% file: relatedWork.tex
\section{Related Works}
\label{sec:related}

Compared to traditional VI, neural density estimators offer a richer family of approximate posterior distributions. Neural density estimators mainly include two families: normalizing flows (NF)~\citep{rezende2015variational} and autoregressive flows (AF)~\citep{Larochelle2011,uria2016neural}. In the former group of methods, the goal is to find an invertible function that transforms a random variable drawn from a base density (\eg a standard Gaussian) to a target density. In addition to the invertibility, the function should have a tractable {$\log$ determinant of the Jacobian}. In the latter group, the target density is modeled as the product of conditional densities. Several works drew connections between the two families~\citep{kingma2016improved,papamakarios2017masked,huang2018neural}.   

To model a density, $p(\vx)$, of a random variable, $\vx$, in a high dimensional space, an AF first assumes an ordering between coordinates of the variable and models the conditional distribution of $x_{i}$, given the previous coordinates, $\vx_{1:i-1}$, as $p(\vx) = \prod_{i}{p(x|\vx_{1:i-1})}$. This recursive formulation can be modeled by a recurrent architecture~\citep{uria2013rnade} where the conditional distributions are assumed to be a function of a hidden state. The dependency on the ordering of the random variables is one of the drawbacks of the AF (see~\cite{papamakarios2017masked} for an illustration) and several methods are proposed to alleviate the problem~\citep{germain2015made,papamakarios2017masked}. Also, in a high dimensional space, generating samples can be expensive~\citep{Oord2016}. Unlike AFs, our method is not dependent on an ordering,  and sampling is relatively inexpensive. Our architecture uses an RNN and may have some resemblance with the AF methods, but the RNN in our method approximates the integration of an ODE and not a conditional distribution. 

Our approach is closer to the NF. An NF method starts with random draws from a known distribution and applies a chain of invertible transformations $f_t$,
\begin{IEEEeqnarray*}{ccc}
  \vz_0 \sim q(\vz_0 | \vx), &\quad& \vz_t = f_t (\vz_{t-1},\vx).
\end{IEEEeqnarray*}   
The main challenge is to ensure that $f_{t}$ is invertible and the determinant of its {Jacobian} can be efficiently computed. NFs were first introduced by~\cite{rezende2015variational}. They proposed a planar transformation, where $f_{t}(\vz_{t-1}) = \vz_{t-1} + \vu h(\vw^T \vz_{t-1} + b)$ and $h(\cdot)$ is a non-linearity. $f_{t}$ is invertible under some constraints (see the Appendix of \cite{rezende2015variational} for more details). However, the planar family is limited to an MLP with a single node bottleneck layer. \cite{tomczak2016improving} and \cite{van2018sylvester} proposed models belonging to a volume preserving family. The volume preserving family has limitations in modeling multi-modal densities.

\cite{papamakarios2017masked} and \cite{kingma2016improved} drew connections between NF and AF. To keep the computation of {$\log$ determinant of Jacobian} tractable, they use an affine form between $\vz_{t-1}$ and $\vz_{t}$ which results in a lower triangular Jacobian whose determinant is very cheap to compute. \cite{huang2018neural} recently proposed  Neural Autoregressive Flow (NAF) which extends the previous two papers by replacing the affine transformation with a more rich family of transformation. They ensures the invertibility of the flow by using a monotonic function on the bottleneck layer. Although the resulting flow is invertible, to the best of our knowledge, computing the inverse is not straightforward. Alternatively, we do not have any constraint on the architecture of the MLP and we are able to invert the flow simply by integrating the velocity field backward.

Very recently, \cite{NeuralODE2018} proposed a continuous time RNN based on an ODE solver which can be used as a building block for NF. While we use the Euler method, they propose to use the Runge–Kutta method~\citep{Dormand1980} which results in more accurate integration. Integrating their method with ours can improve the performance of our model, but we will leave this for future work. Finally, there are several works at the intersection of deep learning and ODE/PDE~\citep{long2017pde,Haber2018} that are tangentially related to our work, and their proposed methods can potentially improve the performance of our diffeomorphic flow.

%% file: experiments2.tex

\section{Experimental Results}

In this section, we evaluate the proposed method in four different experiments: (1) \textbf{Forward and Backward Flows}, in which we evaluate the accuracy of the forward and backward passes of the flows in approximating the solution of an ODE defined in Section~\ref{eq:ODE}. (2) \textbf{Expressiveness of \DNF}, where we study the expressiveness power of the our method on two toy distributions and compare the results with the state-of-the-art and traditional sampling techniques (\eg MCMC). (3) \textbf{Effect of Regularization}, where we study the effect of the regularizers introduced in Section~\ref{sec:regularizing_the_flow} on our flow. (4) \textbf{Variational auto-encoder on MNIST}, where we apply our approach on the MNIST dataset~\citep{MNISTdataset} using the VAE application and compare the results with state-of-the-art methods.
In all of these experiments, $T$ denotes the number of ResNet cells in the stationary velocity block (Figure~\ref{fig:general}-Middle shows one such stationary velocity block), and $K$ denotes the number of non-stationary velocity blocks in our flow (Figure~\ref{fig:general}-Top). The total number of cells in the flow is denoted $N=K\times T$.


\begin{figure}
    \centering
    \includegraphics[width=1.0\linewidth]{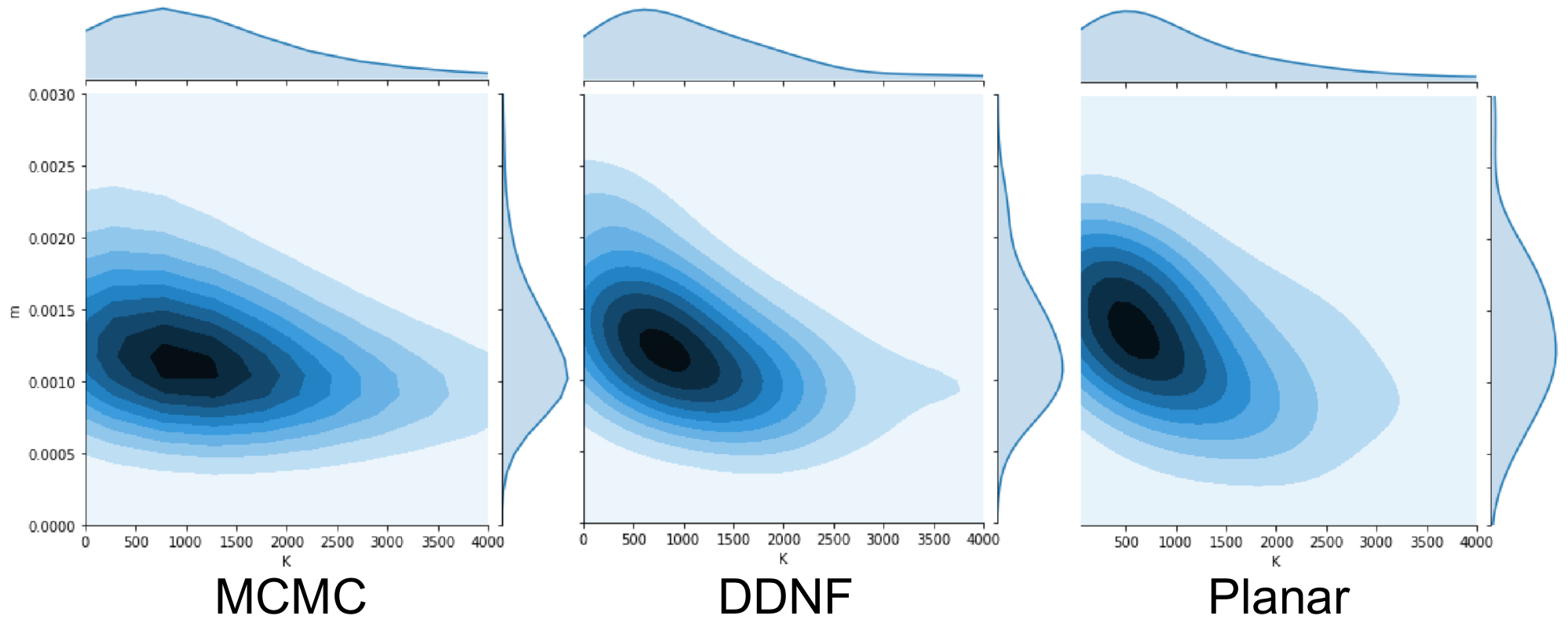}
    \vspace{-0.5cm}
    \caption{\small Comparing the posterior estimation of the over-dispersion model showing one instance out of 10 repetitions of the experiment (see Appedix~B  for quantitative results and details).}
    \label{fig:skwP}
\end{figure}

\subsection{Forward and Backward Flows}
\begin{figure}[t]%
    \centering
     \includegraphics[height=2.8cm]{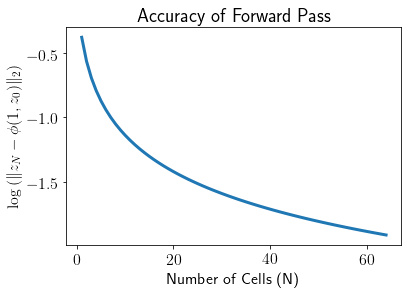}
     \includegraphics[height=2.8cm]{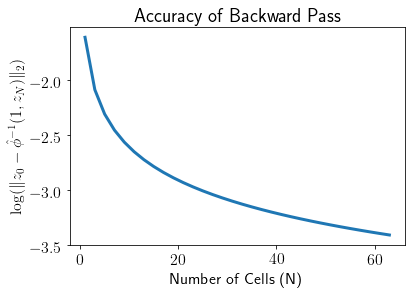}
     \caption{\small 
     Comparing the accuracies of the Forward and Backward passes Diffeomorphic flow. We use $\hat{\phi}$ and $\phi$ for our the highly accurate \texttt{ode45}, respectively. 
    (Left) Comparing the accuracy of the forward pass of the Diffeomorphic flow ($\hat{\phi}(1,\vz)$) with \texttt{ode45} ($\phi(1,\vz)$).
    (Right) Comparing the accuracy of the backward pass in recovering the input, \ie $ \vz_0 \approx \hat{\phi}(1,\vz_K)$.
    }
    \label{fig:ode_solver}%
\end{figure}

We perform the following experiments to evaluate the accuracy of our method in the forward pass (\ie$\phi(1,\vz)$) and backward pass (\ie$\phi^{-1}(1,\vz)$). We construct a randomly initialized \DNF~ flow which takes as input $\vz_0 \in \mathbb{R}^{2}$ and transforms it into $\vz_K \in \mathbb{R}^{2}$. We experiment with various number of ResNet cells $T$ in the stationary velocity block. The velocity field $\vv_k(\cdot)$ in each ResNet cell (see Figure~\ref{fig:general}-Top) is parameterized by two fully connected layers with two hidden units each. We randomly draw samples, namely $\vz_0 \sim \gN(0,\rmI)$, and pass the samples through the flow, the output of  which is $\vz_{K}$. We use \texttt{ode45}\footnote{ We use the  \textit{scipy.integrate.ode} function.} \citep{hairer1993solving} to compute highly accurate forward integration as the \emph{ground-truth} (\ie $\phi(1,\vz)$). \figurename~{\ref{fig:ode_solver}}-left reports the mean squared error between the solution of our flow and that of the ode45 solver.

To evaluate the backward direction, we pass the output of the forward pass, $\vz_{K}$, through our inverse flow (see Figure~\ref{fig:invConst}) and compute the mean squared error of the reconstruction averaged over 50 experiments with different random seeds. We report the findings in \figurename~\ref{fig:ode_solver}-right. As expected, the accuracy increases with the number of cells T of each non-stationary block k. Although our integration scheme is not as accurate as \texttt{ode45}, the backward pass is able to recover the  original latent variable $\vz_{0}$ accurately.

\subsection{Expressiveness of \DNF}\label{sec:expressiveness}
We perform two experiments to show the expressive power of our method. The first is a \textbf{toy energy fitting} experiment following~\cite{rezende2015variational} in which we approximate a set of  $2D$ unnormalized densities $p(\vz) \propto \exp \left[ -U(\vz) \right]$. These densities are chosen to be multi-modal which are hard to capture by typically-used methods such as mean field. The second experiment is a \textbf{posterior estimation } experiment in which we demonstrate the power of our method in approximating the real posterior density of a hierarchical model defined on real data. The setup is adopted from~\cite{Salimans2013}.

\paragraph{Toy energy fitting:}
Following~\cite{rezende2015variational}, we use two unnormalized densities shown in the first column of Figure~\ref{fig:toyDensity} (for the expression of these densities, please check the Appendix~A).

We applied three different flows on the initial distribution ($q_{0}(\vz) = \gN(0,I)$): \DNF, the planar NF ~\citep{rezende2015variational}, and the neural autoregressive flow (NAF)~\citep{huang2018neural}. We experimented with varying number of flows (non-stationary velocity blocks in our case) $K \in \{2, 8, 10\}$. All of the methods performed well for $K=10$, but our method and NAF achieve high quality approximation of the density with less number of flows (\eg $K=2$). Note that NAF enjoy a richer parametrization while our velocity field is a simple two layer neural network with two hidden units each. We were not able reduce the number of parameters in the NAF without compromising its performance.

\paragraph{Posterior estimation:}
In this section, we consider a hierarchical model for estimating stomach cancer rates of a few large cities in Missouri. The model is originally introduced in~\cite{Albert2009} and also studies in~\cite{Salimans2013}. The data consists of 20 pairs, $(n_j, y_j)$, where $n_j$ and $y_{j}$ denote number of individuals at risk and the number of cancer deaths respectively. \cite{Albert2009} proposed the beta-binomial to model observation pairs and an improper prior for the parameters of the beta-binomial, $m$ and $L$. The hierarchical model can be written as follows,
\begin{IEEEeqnarray*}{l}
    p(m,L) \propto \frac{1}{m(1 - m)(1+L)^2}, \\
    p(y_j|m,L) = {n_j \choose y_j} \frac{B(Lm+y_j,L(1-m)+n_j - y_j)}{B(Lm,L(1-m))},
\end{IEEEeqnarray*}
where $B(\cdot,\cdot)$ is the Beta-function. The results are shown in \figurename~{\ref{fig:skwP}} for MCMC, \DNF~, and planar-NF~\citep{rezende2015variational}. We note here that we were not able to get NAF to converge on this dataset. We view the MCMC density as the ground-truth. The marginal posterior distributions demonstrate that our method results in a closer approximation of the correct posterior. For more analysis and quantitative comparison see Appendix~B.

\subsection{Effect of Regularization}

We perform experiments to investigate the two regularization schemes introduced in Section~\ref{sec:regularizing_the_flow}.  For the regularization experiments, we set $K=8$. We also use the first toy density, which is defined in Section~\ref{sec:expressiveness}, to study the effect of regularization on our flow.

\paragraph{Velocity field regularization:}
We noticed that the flow seems to be stable for sufficient $T$. However, when the dimensionality of the problem is high, one may prefer to reduce the computational cost by reducing $T$ or coarsen the approximation of the $\log \det (\cdot)$ in~\eqref{eq:taylor} by ignoring the term $(\Delta t)^{2}$. This could cause an instability in training. To mitigate this, we define a regularized ELBO,
$\max_{\phi} \left(\gF(\phi) - \gamma \Gamma(\phi)\right)$,
where $\gF$ denotes the ELBO as a function of the flow, $\Gamma(\phi)$ is a velocity field regularizer defined in~\eqref{eq:geodes}, and $\gamma \in \mathbb{R}$ is a hyper parameter (a regularization constant).  

\figurename~{\ref{fig:stability}} shows the results for the first and second order approximation of the $\log \det$ function with regularization ($\gamma=0.1$) and without regularization ($\gamma=0$). The variance of the ELBO is reported as a measure of stability of the optimization. To show the flow's effect, we apply it on a regular grid (see Appendix~A for an example) and visualize how the flow \emph{deforms} the grid. We also show the displacement field (\ie $\Delta \phi := \phi -id$) which shows the start and end location of sampled particles in a given domain to which the flow is applied. Notice, ignoring the second order term of the $\log \det (\cdot)$ results in worse numerical instability. The twist in the grid suggest that the flow is not invertible. Adding the second order term improves the stability but the flow is highly irregular. Adding the regularization stabilizes both approximations due to velocity shrinkage that makes the Taylor expansion more accurate. Hence, the regularization helps achieving a smooth flow even in extreme case of $T=1$ and the first-order Taylor expansion.

\begin{figure}
   \centering
       \includegraphics[height=3.cm]{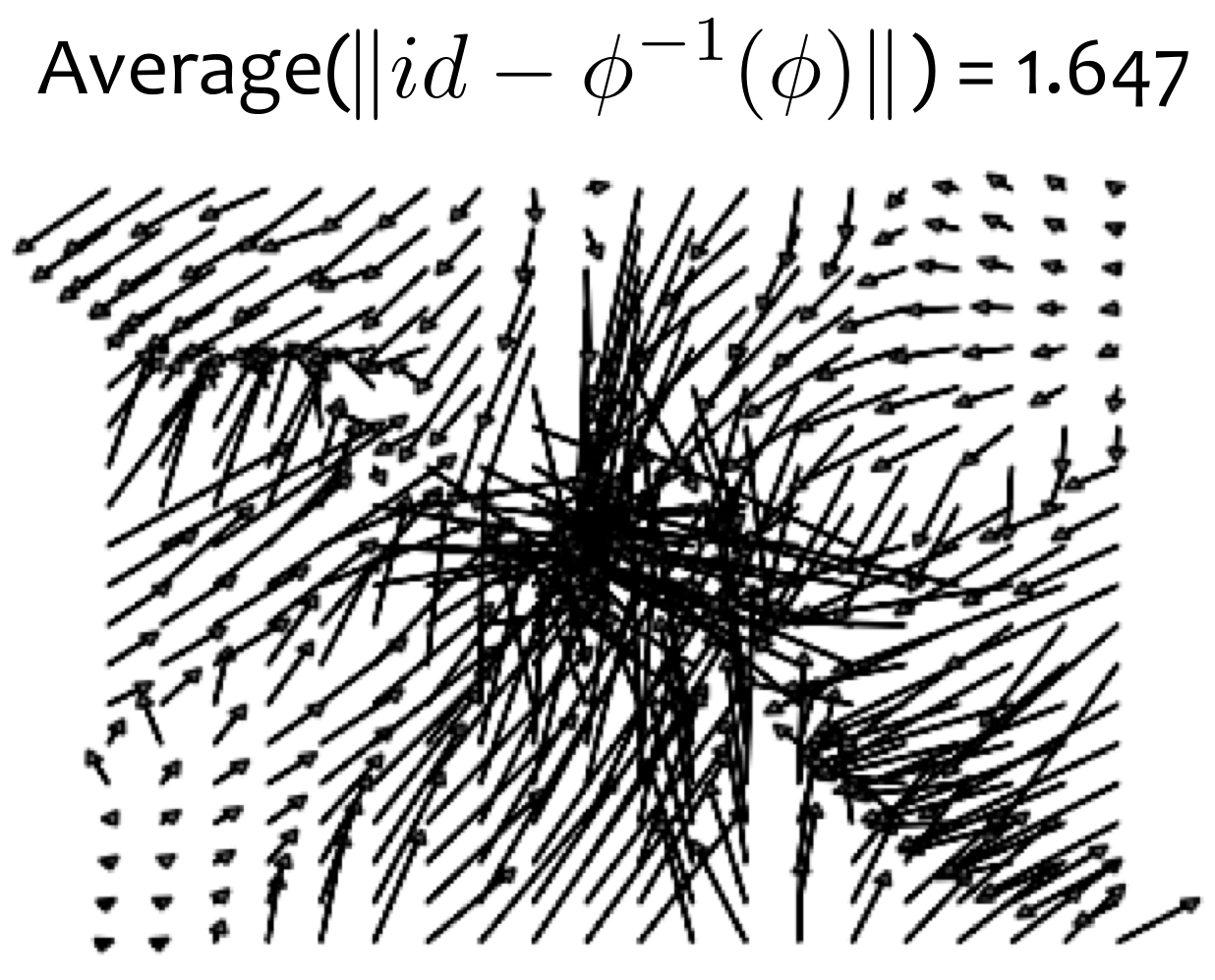}
       \includegraphics[height=3.0cm]{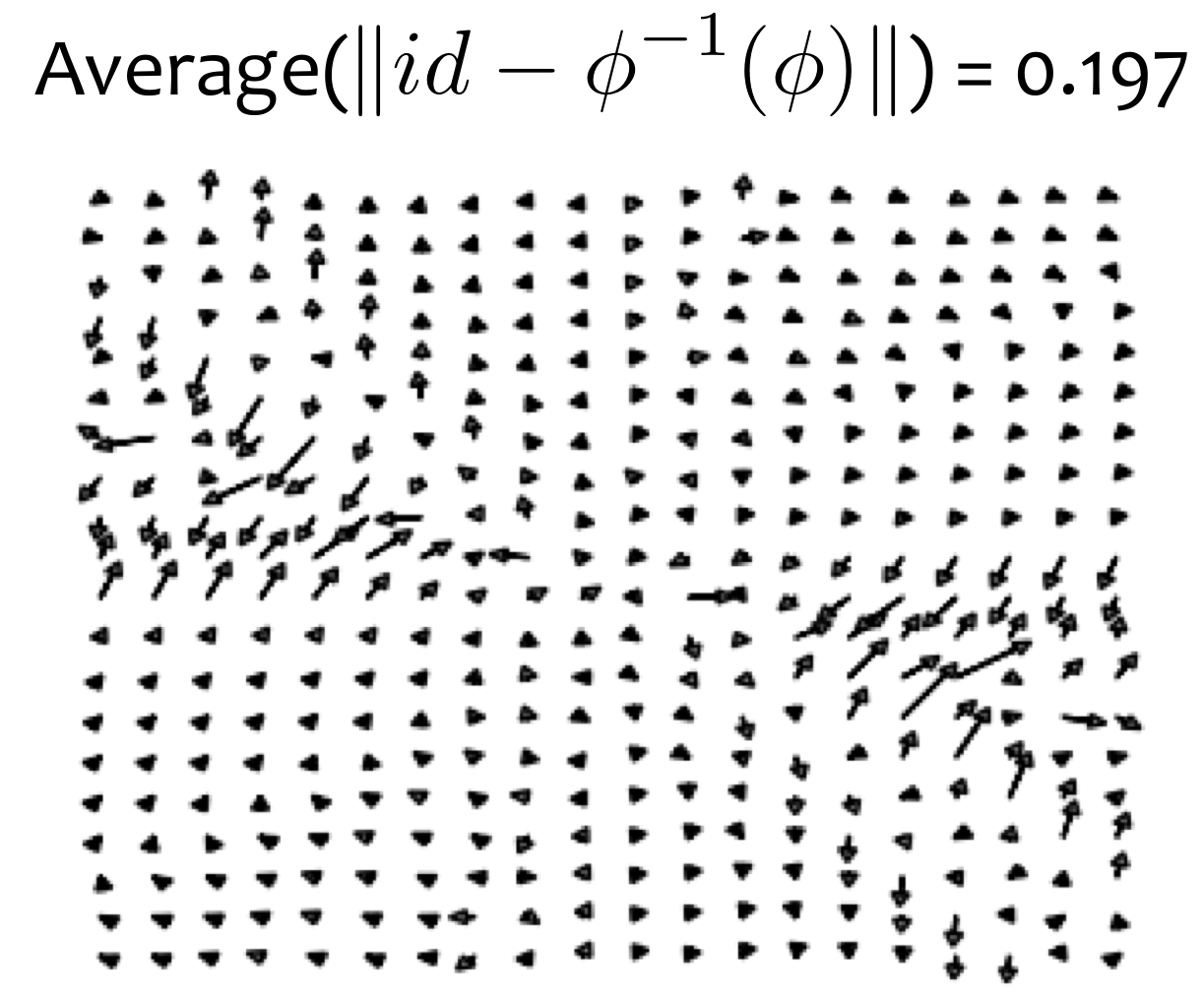}
    \vspace{-0.3cm}   
    \caption{\small
    The effect of inverse consistency on the invertibility of the flow when the ODE is coarsely approximated ($T=1$). The figures visualizes the displacement with of the flow composed with its inverse, \ie $id - \phi^{-1}(\phi)$. The regularization significantly improves the invertibility. }
    \label{fig:invConstexp}
\end{figure}

\vspace{-.2cm}
\paragraph{Inverse consistency:}
We performed a similar experiment with the inverse consistency regularizer. Again, we observed that the inversion of a flow is of high quality when $T$ is sufficiently large whereas a small $T$ results in a coarse approximation of the ODE hence compromising the quality of \textbf{invertibility}. We setup an experiment where we set $T=1$ and $K = 8$, and we optimize the regularized ELBO, \ie $\max_{\phi} \left(\gF(\phi) - \gamma \gR(\phi)\right)$,
where $\mathcal{R}(\phi)$ is an inverse consistency regularizer as defined in~\eqref{eq:invConst}. \figurename~{\ref{fig:invConstexp}} reports the displacement field for the composition of the flow and its inverse, namely $ \left(id -  \phi^{-1}(1,\phi{1,\cdot})\right) $. It also report the average L2 norm of this displacement field namely, $average(\| id -  \phi^{-1}(1,\phi{1,\cdot}) \|_2)$; ideally, this value should be zero. Figure~\ref{fig:invConstexp} shows that adding the regularizer improves the invertibility even in the extreme case of $T=1$. Hence, if the computaional cost and invertibility are concerning, one can reduce $T$ and add the inverse consistency regularizer.

\subsection{Variational auto-encoder on MNIST}
We evaluate the \DNF's ability to improve variational inference. We run variational autoencoder (VAE) experiments on the MNIST ~\citep{MNISTdataset} and Omniglot~\citep{lake2015human} datasets, and we compare the performance of our model against three models from the literature: vanilla VAE, planar Normalizing Flows (NF), and Neural Autoregressive Flows (NAF). We run two variants of our method one without context (\ie $q_0(\vz)$ is standard Gaussian) and with context, \ie $q_0(\vz| \vx)$ receives input from the encoder similar to NAF. The results  are reported in Table~\ref{table:MNIST_Omniglot_comparisons}. \DNF~outperforms the vanilla VAE and the NF model by a statistically significant margin. A context aware variant of our method (\ie a signal from the encoder is fed to the flow in addition to the sampled noise) produces a comparable results to NAF (which also uses such signal) highlighting the importance of the context from the encoder. See Appendix~C for more details of this experiment. 

\begin{table}[t]
\centering
\caption{\small
Comparison of the negative ELBO on the test sets of the  MNIST and Omniglot datasets with $\dim(\vz)=40$. The reported results are averaged over three experiments. [1]: \citep{rezende2015variational}, [2]: \citep{huang2018neural}.}
\label{table:MNIST_Omniglot_comparisons}
\small
\begin{tabular}{l|c|c}
Model  &  MNIST & Omniglot \\ \hline \hline
Vanilla VAE  & 	103.46 $\pm$ 0.49 & 124.32$\pm$ 0.09\\
planar-NF [1] &  102.14 $\pm$  0.23 & 124.13$\pm$ 0.02\\
NAF [2] & 90.31 $\pm$ 0.13 &  110.46 $\pm$  0.08\\
\DNF~ &  101.14 $\pm$ 0.32 &  121.18 $\pm$  0.23\\
\textbf{\DNF~ + context} &  \textbf{88.97 $\pm$ 0.56} & \textbf{109.01$\pm$ 0.06}\\
\hline
\end{tabular}
\end{table}

%% file: conclusion.tex
\section{Conclusion}
\label{sec:diss}
In this work, we developed a new type of NF for density estimation. Our invertible flow consists of compositions of many tiny mappings. Jacobians of the almost identity-like small mappings were approximated using Taylor expansion which is essential to control the computational cost of the algorithm. Such construction mimics Euler discretization of an ODE governed by a vector (velocity) field modeled by an MLP. In contrast to previous works, we have no limitation on the architecture of the MLP (except smoothness), yet we are able to invert the flow accurately. We believe the close connection of this work with Riemannian Geometry and Lie Algebra can potentially open new direction of research for neural density estimation in the future.   

%% file: appendix.tex
\large{\section*{Deep Diffeomorphic Normalizing Flows}}

\appendix
\appendixpage
\addappheadtotoc
\begin{appendices}

\section*{A~~Toy Densities}\label{appen:toy-density}
\begin{figure}[h]
    \centering
    \includegraphics[width=1.0\linewidth]{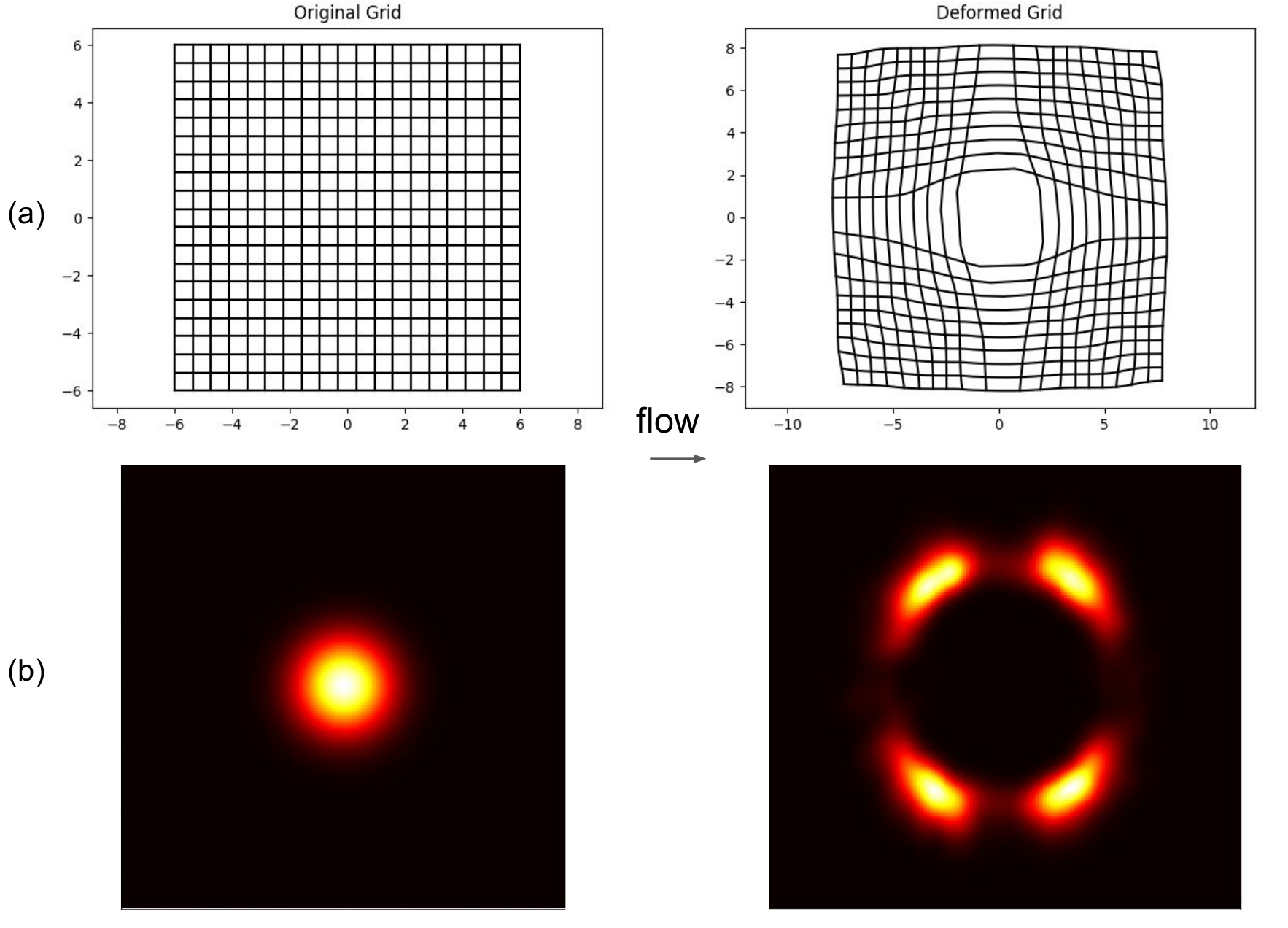}
    \caption{The result of apply our \DNF flow on (a) a regular grid, and (b) on a uni-model Gaussian distribution. }
    \label{fig:illustration_grid}
\end{figure}

The toy densities used in Section~\ref{sec:expressiveness} are deined as follows,
\begin{itemize}
   \item $\scriptstyle U(\vz) = \tfrac{1}{2}\left( \tfrac{\| \vz \|^{2}_{2} - 4}{0.4} \right)^{2} -\ln \left( e^{-\frac{1}{2} \left[ \tfrac{z_1 - 2}{0.8} \right]^2} + e^{-\tfrac{1}{2}\left[ \tfrac{z_1 + 2}{0.8} \right]^2 } \right)$
   \item $\scriptstyle U(\vz) = \tfrac{1}{2} \left( \tfrac{\| \vz \|^2_2 - 2}{0.4} \right)^2 -\ln \left( e^{-\frac{1}{2}\left[ \tfrac{z_1 - 2}{0.8} \right]^2} + e^{-\tfrac{1}{2} \left[ \tfrac{z_1 + 2}{0.8} \right]^2}  \right)$
\end{itemize}
Figure~\ref{fig:illustration_grid} shows the effect of applying our \DNF flow to a regular grid and and a uni-modal Gaussian distribution.

\section*{B~~Posterior Approximation}
\begin{figure*}[t]
   \centering
   \includegraphics[width=.8\linewidth]{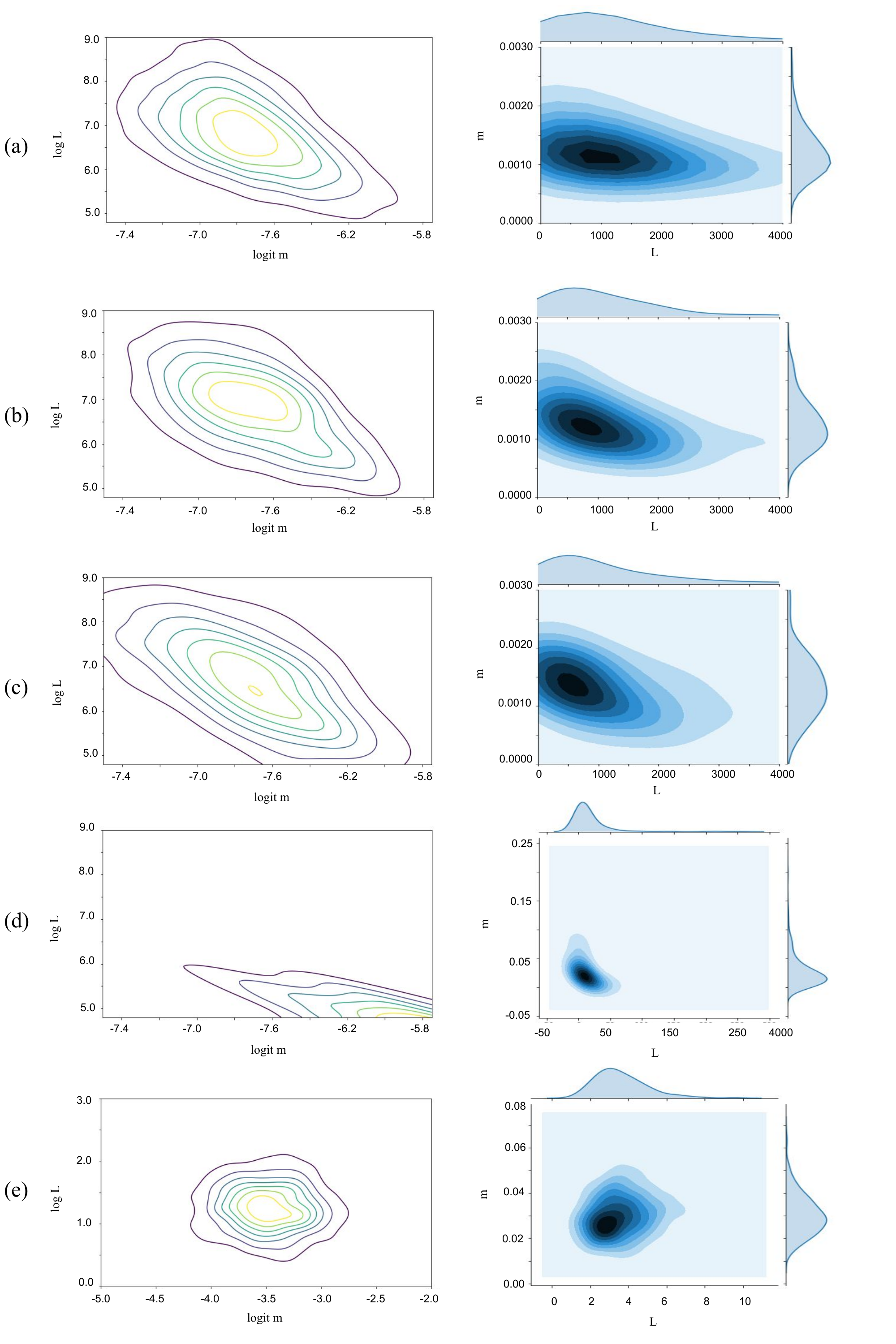}
   \caption{Comparing the posterior approximation of the over-dispersion for different method: (a) MCMC, (b) \DNF, (c) planar-NF, (d) Householder, (e) ADVI.  Adding the scalar and drift transformations to the planar mapping improves the results. Neither Householder, nor ADVI were able to capture the true posterior (MCMC).}
   \label{fig:skewed_posterior_appendix}
\end{figure*}

We report the performance of other posterior estimation techniques than what we showed in Section~\ref{sec:expressiveness}. Other methods include Automatic Differentiation Variational Inference (ADVI)~\citep{kucukelbir2017automatic} and Householder flow (HF)~\citep{tomczak2016improving}). For both of these methods, we used their implementation in the PyMC$_3$ package (https://docs.pymc.io/).  We repeat the experiment 10 times and the results of one instance is show in Figure~\ref{fig:skewed_posterior_appendix}.  \DNF approximates the true posterior(MCMC) accurately. Adding the scalar and drift transformations to the planar mapping improves the accuracy. The Table 2 reports the posterior mean of the two variables for MCMC, \texttt{Plana + scalar + drift}, and \DNF. 

\begin{table}[]
\caption{\small
Comparison of the average estimated $m$ and $L$ for the posterior experiment using MCMC, \DNF, and planar-NF. As can be seen, the estimated values by our flow are closer to those estimated by MCMC than the planar-Flow. The results are averaged over ten independent experiments.}
\begin{tabular}{l|l|l}
          & m  ($\times {10}^{-6}$)       & L                               \\ \hline
MCMC      & 1290.0 $\pm$ 6.5   & 1560.9 $\pm$ 65.0  \\ \hline
DDNF      & 1289.7 $\pm$ 25.1  & 1354.9 $\pm$ 71.54 \\ \hline
planar-NF & 1600.0 $\pm$ 594.7 & 1268.1 $\pm$ 390.8
\end{tabular}
\end{table}

\section*{C~~Training Setup for the VAE Experiment}
\paragraph{Network Architecture:}
For the MNIST dataset, we implement the encoder of the VAE as an MLP with one hidden layer of size 128 and a latent code of dimension 40. The decoder is also an MLP with one hidden layer of size 128 that takes in an input of size 40 and outputs a vector of size 28$\times$28.  The velocity fields in the \DNF~ are parameterized by two hidden layer with two hidden units each. We use tanh activation across all the hidden layers, and we use sigmoid activation on the output of the VAE network.
\paragraph{Training Details:}
We train using SGD implemented in tensorflow. Across all experiments, we use a batch size of 100 and a learning rate of 0.001 and we train for 400 epochs. We report the minimum -ELBO on the test set for each of the methods averaged over three experiments with different random seeds (but common accross experiments).

\paragraph{Low Dimensional Latent Space:}
We test our method on a lower dimensional latent space setting of the VAE where we use the same setup discussed in the previous sections, but with a latent code of dimension 2. Table~3 shows the results. 

\begin{table}[t]
\centering
\caption{\small
Comparison of the ELBO on the test set for the  MNIST dataset ($\dim(\vz)=2$). The reported results are averaged over three experiments. [1]: \citep{rezende2015variational}, [2]: \citep{huang2018neural}.}
\label{table:MNIST_comparisons}
\small
\begin{tabular}{l|c}
Model  &  -ELBO\\ \hline \hline
Vanilla VAE (diagonal covariance) & 	149.43 $\pm$ 0.14\\
planar-NF [1] &  148.97 $\pm$  0.29\\
NAF [2] & 144.10 $\pm$ 0.50\\
\DNF~ &  147.27 $\pm$ 0.58\\
\textbf{\DNF~ + context} &  \textbf{143.68 $\pm$ 0.51}\\
\hline
\end{tabular}
\end{table}

\section*{D~~Deriving the Determinant of the Jacobian}

\begin{IEEEeqnarray*}{l}
\log \det(\mathcal{J} \phi^{\vv}(\Delta t,\cdot)) = \log \det  \underbrace{ \left( I+\Delta t \mathcal{J} \vv(\cdot) \right)}_{A} \\
\end{IEEEeqnarray*}
\vspace{-1.2cm}
\begin{IEEEeqnarray*}{l}
    =\tfrac{1}{2} \log \det ( I + \\
    \Delta t \underbrace{ \left( \gJ \vv(\cdot) +  \gJ \vv(\cdot)^T + \Delta t \gJ \vv(\cdot)^T \gJ \vv(\cdot) \right)}_{B} ) 
\end{IEEEeqnarray*}
\vspace{-0.5cm}
\begin{IEEEeqnarray*}{ll}
    \approx & \tfrac{1}{2}\Delta t \text{Tr}(\gJ \vv(\cdot) + \gJ \vv(\cdot)^T)  \\  
            & -\tfrac{1}{2}(\Delta t)^2 \text{Tr}(\gJ \vv(\cdot)^T \gJ \vv(\cdot)) 
\end{IEEEeqnarray*}

where the first equality follows from the architecture of the ResNet cell in \figurename~{\ref{fig:general}}-Top, the second equality is a results of $\det(A)^2 = \det(AA^T) $, and the approximation is the second order Taylor expansion of $\log \det (I + \Delta t B)$ around $B = 0$. We ignore polynomials terms of $\Delta t$ with the degree of three and higher.

\section*{E~~Other Choices of the Hilbert Norm}
Several choices are possible of the Hilbert norm. 

\paragraph{Identity and Laplacian operator}:
Inspired by the research in the medical imaging community~\cite{Beg2005,Zhang2015}, we set $\rmL = (\alpha \Delta_\vz + \rmI)^{c}$ where $c$ is an integer power and $\Delta_{\vz}$ is the Laplacian operator with respect to $z$. Since the Laplacian is a negative semidefinite operator, $\alpha \leq 0$. The Laplacian encourages smoothness of the velocity field, \ie non-smooth velocity field result in large Laplacian value.

There are several advantages of this choice of inner product. For sufficiently large powers, $c$, the geodesic regularization guarantees existence of diffeomorphic flows, as long as the norm of the velocities are bounded. Without this condition, the flow of a differentiable velocity field is only guaranteed to exist over some unknown interval $t \in [0, \epsilon)$, not necessarily up to $t = 1$. This is due to the Picard-Lindel\"of existence theorem for ODEs. However, as shown in \cite{Dupuis1998}, we can guarantee that the flow of a velocity field, given by \eqref{eq:ODE}, generates a diffeomorphism at time $t = 1$, if the space of velocity fields satisfy certain regularity conditions. This regularity condition is that the space of velocities, $V$, be continuously embedded in the Sobolev space $W^{1,\infty}(\Omega, \mathbb{R}^d)$, \ie the space of velocity fields with bounded generalized derivative. 

 Note that the Laplacian operator can be viewed as the trace of the Hessian matrix with respect to $\vz$ and the same trick applied in~\eq~\ref{eq:rndTrace} can be used to efficiently compute the trace. Thanks to the reverse-mode differentiation introduced by~\cite{Pearlmutter1994}, the Hessian-vector products can be computed efficiently in ($\gO(d)$).
If $\alpha=0$, we retrieve the $\ell_{2}$-norm. The $\ell_{2}$ simply shrinks the velocity toward zero. When the \DNF~ has few cells (\ie $\Delta t$ is large), we found that even a simple $\ell_{2}$ regularization of $v$ helps the Taylor expansion in~\eqref{eq:taylor} to be more accurate. In this paper, we set the $\alpha=0$ because we use $tanh(.)$ for non-linearity and our velocity field is smooth by construction.


\end{appendices}